# Neural Text Generation: Past, Present and Beyond

Sidi Lu[1], Yaoming Zhu[1], Weinan Zhang[1], Jun Wang[2], Yong Yu[1]
[1]Shanghai Jiao Tong University, [2]University College London
{steve_lu, ymzhu, wnzhang, yyu}@apex.sjtu.edu.cn, j.wang@cs.ucl.ac.uk

## Abstract

This paper presents a systematic survey on recent development of neural text generation models. Specifically, we start from recurrent neural network language models with the traditional maximum likelihood estimation training scheme and point out its shortcoming for text generation. We thus introduce the recently proposed methods for text generation based on reinforcement learning, re-parametrization tricks and generative adversarial nets (GAN) techniques. We compare different properties of these models and the corresponding techniques to handle their common problems such as gradient vanishing and generation diversity. Finally, we conduct a benchmarking experiment with different types of neural text generation models on two well-known datasets and discuss the empirical results along with the aforementioned model properties.

## 1 Introduction

Natural language processing (NLP) problems, especially natural language generation (NLG), have long been considered as among the most challenging computational tasks [Murty and Kabadi, 1987]. NLG techniques are widely adopted as the critical module in various tasks, including the control-free sentence or poem generation [Zhang and Lapata, 2014] and input-conditioned text generation such as image captioning [Karpathy and Fei-Fei, 2015] and sentiment/tense controlled sentence generation [Hu et al., 2017] etc.

The problem is challenging due to a few reasons. Generally speaking, there is information imbalance between the input and output in these tasks [Shapiro, 1992] especially for the cases with non-text input. The semantic of input is usually specified and clear, whereas components of natural language are often ambiguous. This fact forces the neural text generation (NTG) models to find the common patterns of the target language and express the input information with disambiguation through constructing appropriate contexts. During this procedure, there are mainly two difficulties. One is the grammatical complexity of natural language[1]. The other is about the difficulties during the extraction, simplification and transformation of the input information. The latter one is too task-specific, thus is not the focus in this paper. To address the grammatical problems, researchers develop general approaches to build complicated knowledge-based systems as is discussed in [Reiter and Dale, 2000]. It is important to point out that although this paradigm needs much human effort, it is still widely used in many commercial products today since it is interpretable and robust if well designed.

During the recent decade, neural network (NN) and its variants have shown promising results in many tasks. For text generation, the neural network language model (NNLM) [Bengio et al., 2003] is first proposed to exploit the advantages of NN for text generation tasks. NNLM can be regarded as a direct extension of the n-gram paradigm with the generalization ability of NNs. Given the ground truth sequence $s_n = [x_0, x_1, ..., x_{n-1}]$ and a $\theta$-parametrized language model $G_\theta(x|\text{context}) = \hat{P}(x|\text{context})$ (similarly hereinafter), a typical NNLM adopts an approximation as

$$\hat{P}(x_t|\text{context}) \approx P(x_t|x_{t-n+1}, x_{t-n+2}, ..., x_{t-1}) . \quad (1)$$

However, the n-gram paradigm is theoretically impossible to capture long-term dependencies, according to some previous criticism [Rosenfeld, 2000]. To address this problem, recurrent neural network language model (RNNLM) [Mikolov et al., 2010] is developed, which is a more general implementation for a language model with Markov property. A typical RNNLM uses a recurrent neural network (RNN) to auto-regressively encode previous variant-length inputs into a "hidden" vector, which is then used during the inference of the next token. This procedure can be formulated as

$$\hat{P}(x_t|\text{context}) \approx P(x_t|\text{RNN}(x_0, x_1, ..., x_{t-1})) . \quad (2)$$

Neural text generation has been well studied since RNN-LM. For example, by adopting improved variants of RNN, e.g., long short-term memory (LSTM) [Hochreiter and Schmidhuber, 1997] and gated recurrent unit (GRU) [Cho et al., 2014], RNNLMs show acceptable results in capturing long-term dependencies in the text. However, as pointed out by [Bengio et al., 2015], fitting the distribution of observed data does not mean generating satisfactory text because of the *exposure bias*. Later, various kinds of solutions are proposed, including reinforcement learning (RL) based models, generative adversarial nets (GANs) [Goodfellow *et al.*, 2014; Yu *et al.*, 2017] frameworks and end-to-end re-parameterization techniques [Kusner and Hernández-Lobato, 2016].

---

[1]Most natural languages are recursively enumerable languages a.k.a. Type-0 languages according to Chomsky Hierarchy [Chomsky, 1956]. This is the most complicated type which requires a Turing Machine to handle it.

In this paper, we present a systematic survey on these recently proposed neural text generation (NTG) models. We carefully discuss different properties of these models and the corresponding techniques to handle their common problems such as gradient vanishing during training and generation diversity. Compared to a previous work [Xie, 2017] that is mainly on sequence-to-sequence (Seq2Seq) models, this paper focuses more on recently proposed RL and GAN based methods while Seq2Seq is a special case of the basic MLE methods. Finally, we conduct a benchmarking experiment with different types of neural text generation models on two well-known datasets and discuss the empirical results along with the aforementioned model properties. We hope this paper could provide with useful directions for further researches in this field.

## 2 On Training Paradigms of RNNLMs

In this section, with a combination of chronological and technique categorized presentation, we mainly introduce three training paradigms of RNNLMs, namely supervised learning, reinforcement learning techniques and adversarial training schemes.

### 2.1 NTG with Supervised Learning

Although text generation is actually an unsupervised learning task, there do exist some supervised metric that are good approximations of the ground truth under some constraints. These algorithms focus on directly optimizing some supervised metric. Some of them may include some useful tricks to help in alleviating some specific problems.

**Maximum Likelihood Estimation**
Typically, classical neural language models are trained through maximum likelihood estimation (MLE, a.k.a. teacher forcing) [Williams and Zipser, 1989]. MLE is natural for RNNLMs, since it regards the generation problem as a sequential multi-label classification and then directly optimizes the multi-label cross entropy. The objective of a language model $G_\theta$ trained via MLE can be formulated as

$$J_\theta(s_n) = -\sum_{t=0}^{n-1} \log \hat{P}(x_t|s_t) \, , \qquad (3)$$

where $s_0$ represents the empty string.

It is important to note that, up to now, most existing applied state-of-the-art NTG models adopt MLE as their training objective [Karpathy and Fei-Fei, 2015; Hu *et al.*, 2017]. MLE has better convergence rate and training robustness compared to other algorithms.

However, in theory, MLE suffers from so-called *exposure bias* [Huszár, 2015], which is due to the inherent difference between the training stage and inference stage of language models trained via MLE. That is to say, the language model is trained to generate appropriate succeeding token given ground truth prefix during training time. However, during the inference stage of free generation, the model needs to predict the succeeding token given a generated prefix. There is no guarantee that the model will still behave normally in those cases where the prefixs are a little bit different from those in the training data. The effect of exposure bias becomes more obvious and serious as the sequence becomes longer, making MLE less useful when the model is applied to long text generation tasks.

**Scheduled Sampling**
Scheduled Sampling (SS) [Bengio *et al.*, 2015] is proposed to alleviate the problem. It introduces a random variable $\epsilon$ to reconcile teacher forcing and free generation in order to close the difference between training stage and inference stage.

During each step of the SS training procedure, the evaluation result of $\epsilon$ determines whether the model performs teacher forcing or free generation. SS looks reasonable intuitively, and does show observable improvement compared with vanilla MLE. However, it is soon proved to be *inconsistent* [Huszár, 2015].

### 2.2 NTG with Reinforcement Learning

Text generation with RNNLMs can be viewed as a Markov decision process (MDP), the locally optimal policy of which can be found through reinforcement learning (RL) [Sutton and Barto, 1998].

**PG-BLEU**
A straightforward idea is to use RL policy-gradient algorithms (e.g., REINFORCE) [Sutton and others, 1999] to optimize some non-differentiable metrics. A classical choice is to optimize BLEU [Papineni *et al.*, 2002], an n-gram level metric for language model evaluation. This idea is then derived as PG-BLEU. PG-BLEU optimizes BLEU via REINFORCE, whose minimization objective can be formulated as

$$J_\theta(\hat{s}_n) = -\sum_{t=0}^{n-1} R_t \log \hat{P}(x_t|s_t) \, , \qquad (4)$$

where $\hat{s}_n \sim G_\theta(\cdot)$ is the completed sequence sampled from the generator $G_\theta$ and $R_t = \mathbb{E}_{s \sim G_\theta(\cdot|s_{t+1})}[\text{BLEU}(s)]$ is the expected BLEU score given the prefix $s_{t+1}$ and the generation policy $G_\theta$ to follow.

There are some problems about this algorithm. First, BLEU is not a computationally cheap metric, making PG-BLEU barely useful in practice. Second, BLEU is not a perfect metric even not a strong one as it just counts the n-gram statistics similarity between the generated text and the reference text (corpus). Therefore, it introduces much unnecessary bias into the model.

**Rethinking about MLE**
From an RL point of view, MLE can be regarded as off-policy imitation learning with episodes sampled from the replay buffer consists of ground truth data. In each step of the training process, the reward $R_t$ in Eq. (11) is fixed to be 1.0. As imitation learning also helps reduce variance in many RL scenarios, this explains why algorithms like PG-BLEU usually includes pretraining with MLE [Yu *et al.*, 2017].

### 2.3 Better NTG with Adversarial Training

The success of generative adversarial nets (GANs) [Goodfellow *et al.*, 2014] inspires researchers to solve the problem through adversarial training.

**Professor Forcing: Adversarial Training as Regularization**
As an early work, Professor Forcing [Lamb *et al.*, 2016] uses adversarial module as a regularization mechanism that closes the gap between teacher forcing and free generation procedures. Lamb *et al.* claim that this would help in alleviating exposure bias.

Denote the parameter of the generator $G_\theta$ as $\theta$, the parameter of the discriminator $D_\phi$ as $\phi$. Denote the concatenated hidden state and memory cell with input $x_t, s_t$ as $B(x_t, s_t)$. Professor Forcing can be formulated as follows:

$$\text{NLL}(\theta) = \mathbb{E}_{(x_t, s_t) \sim p_{\text{data}}} \big[ - \log \hat{P}(x_t | s_t) \big] \quad (5)$$
$$C_f(\theta | \phi) = \mathbb{E}_{s_t \sim p_{\text{data}}, \hat{x}_t \sim G_\theta(x|s_t)} [-\log(D_\phi(B(\hat{x}_t, s_t)))] \quad (6)$$
$$C_t(\theta | \phi) = \mathbb{E}_{(x_t, s_t) \sim p_{\text{data}}} [-\log(1 - D_\phi(B(x_t, s_t)))] \quad (7)$$
$$J_G(\theta) = \text{NLL}(\theta) + C_f(\theta | \phi) + (\text{optional}) C_t(\theta | \phi) \quad (8)$$
$$J_D(\phi) = \mathbb{E}_{s_t \sim p_{\text{data}}, \hat{x}_t \sim G_\theta(x|s_t)} [-\log(1 - D_\phi(B(\hat{x}_t, s_t)))]$$
$$+ \mathbb{E}_{(x_t, s_t) \sim p_{\text{data}}} [-\log(D_\phi(B(x_t, s_t)))] \ . \quad (9)$$

During the training process, the parameters of model are updated such that Eqs. (8) and (9) are optimized alternately.

**Sequence Generative Adversarial Network: Adversarial Reinforcement Learning**
A more direct and natural approach is Sequence Generative Adversarial Network (SeqGAN) [Yu *et al.*, 2017]. SeqGAN consists of two parts. One is the generator, typically implemented as an RNNLM $G_\theta$ parametrized by $\theta$. The other is the discriminator, which is a binary classifier $D_\phi$ parametrized by $\phi$, which is trained to distinguish generated sequences from the ground truth. SeqGAN uses REINFORCE (policy gradient) to optimize the original GAN objective:

$$\min_\theta \max_\phi \mathbb{E}_{s \sim p_{\text{data}}} [\log(D_\phi(s))] + \mathbb{E}_{s \sim G_\theta(\cdot)} [\log(1 - D_\phi(s))] \quad (10)$$

The objectives of SeqGAN can be formulated as

$$J_\theta(\hat{s}_n) = -\sum_{t=0}^{n-1} R_t \log \hat{P}(x_t | s_t), \quad (11)$$

where $\hat{s}_n \sim G_\theta(\cdot)$ is the completed sequence sampled from the generator just like that in Eq. (11) and the expected reward $R_t = \mathbb{E}_{s \sim G_\theta(\cdot | s_{t+1})} [D_\phi(s)]$ is based on the evaluation of the discriminator. In practice, $R_t$ can be estimated through Monte Carlo search as is described in SeqGAN [Yu *et al.*, 2017]. SeqGAN shows promising empirical results and has a bunch of following work [Guo *et al.*, 2017; Fedus *et al.*, 2018].

**Further Discussion about SeqGAN**
Adversarial learning, which is originated from GAN [Goodfellow *et al.*, 2014], has recently developed into a new paradigm of unsupervised learning. Adversarial learning has shown promising results in many unsupervised learning tasks, such as mutual information elimination [Liu *et al.*, 2017] and super-resolution [Ledig *et al.*, 2016]. Another interesting example, is generative adversarial imitation learning [Ho and Ermon, 2016], which aims to provide an estimation of the environment through observing ground-truth episodes, so that the agent is trained to mimic the behavior patterns of the provided ground truth. Such a process also exists in SeqGAN.

| Ranking List | | Case A (Gradient Vanishes) | Case B (No Gradient Vanishing) |
|---|---|---|---|
| | I have a pen which writes well. | 1.0 | 0.9 |
| | I have a pen. | 0.99999 | 0.6 |
| Generated Samples | I have pen a. | 1e-10 | 0.5 |
| | pen have a I am. | 1e-15 | 0.4 |
| | pen a pen pen I am. | 1e-20 | 0.3 |

Figure 1: Different score lists may lead to the same ranking list.

During adversarial training, the discriminator progressively manages to learn an estimated environment, which is actually a residual of the real environment where current agent cannot perform very well. However, the output value of discriminator does not have any observable physical meaning. One attempt to address this problem is to perform Lipschitz constraints on the discriminator, just like that in Wasserstein GAN [Arjovsky *et al.*, 2017] and its improved variants [Gulrajani *et al.*, 2017].

Despite the advantages, SeqGAN still suffers from two major problems. One is the gradient vanishing problem, which means when the discriminator is trained to be much stronger than the generator, it becomes extremely hard for the generator to have any actual updates since any output instances of the generator will be scored as almost 0. This may cause the training stops too early before it comes to the true convergence or Nash Equilibrium. The other one is the mode collapse problem caused by REINFORCE algorithm, which increases the estimated probability of sampling particular tokens earning high evaluation from the discriminator. As a result, the generator only manages to mimic a limited part of the target distribution, which significantly reduces the diversity of the outputs. These problems motivate further researches on improvement of SeqGAN. Several extended variants are proposed.

**On Alleviation of Vanishing Gradient**
An observable fact is that, different score lists may lead to the same ranking list as illustrated in Figure 1. As the figure shows, although the AUC scores of the two classifiers in case A and B are the same, for generative models that are trained via policy gradient, rewards provided by case B leads to much faster and stable convergence of the model. Intuitively this is odd, since a better trained discriminator should lead to better estimation of the latent distribution of the data instead of the contrary.

To address this problem, there are basically two main types of methods. The first method is to use rescaled scores as the reward signals. Maximum-likelihood augmented discrete generative adversarial nets (MaliGAN) [Che *et al.*, 2017] is a typical work derived from this method. It rescales the score obtained from the discriminator as reward via

$$R(s) = \frac{D(s)}{1 - D(s)} \ . \quad (12)$$

Also, to speed up convergence, MaliGAN includes the reward baseline method, which is to keep a running mean and variance of the calculated reward $R(s)$ and then use them to perform a linear rescale on the reward. This method not only helps alleviate gradient vanishing problem, but also to-some-extent improves on the diversity of the model according to our experiments.

The second method is to replace the binary classification score with a ranking score. Adversarial ranking for language generation [Lin *et al.*, 2017] (RankGAN) proposes an adversarial text generation model whose discriminator is replaced by an adversarial ranker $S_\phi$ that is trained to optimize the pair-wise ranking loss:

$$J_\phi(\cdot) = \mathbb{E}_{s \sim p_{\text{data}}} \left[ \log S_\phi(s|U, C^-) \right] - \mathbb{E}_{s \sim G_\theta} \left[ \log S_\phi(s|U, C^+) \right], \quad (13)$$

where

$$S_\phi(s|U, C) = \frac{\exp(\gamma \alpha(s|u))}{\sum_{s' \in C} \exp(\gamma \alpha(s'|u))}$$
$$\alpha(s|u) = \cos(y_s, y_u),$$

$u \sim U$, $U$ is the *reference* set, $C^+$ is sequences from the dataset, $C^-$ is generated sequence. $\gamma$ is the inverse Boltzmann temperature factor which can be adjusted as a hyperparameter. RankGAN shows promising results in improving the convergence performance of SeqGAN in many cases. However, since it requires extra sampling from the original data, the computational cost of it is also higher than other models.

Inspired by both ideas, researchers propose Bootstrapped Ranking Activation (BRA) [Guo *et al.*, 2017]. BRA does not require modifications on the architecture of SeqGAN. It simply uses the ranking information in each batch to rescale the rewards, which can be formulated as

$$R^t(\cdot) = \sigma\left(\delta \cdot \left(0.5 - \frac{\text{rank}(i)}{B}\right)\right), \quad (14)$$

where rank($i$) denotes the $i$-th sequence's high-to-low ranking in the batch, $\delta$ is the activation smoothness hyperparameter, $B$ is the batch size, $\sigma(\cdot)$ is a non-linear function that reprojects the original equidifferent distribution to make BRA more effective and general. In their experiments in LeakGAN [Guo *et al.*, 2017], researchers pick the sigmoid function as $\sigma(\cdot)$, which has also been practically proved effective in our experiments. BRA does not require much extra computation and is easy to be included in other models, making it a competitive gradient stabilizer.

**On the Enhancement of Diversity**
A special case of the mode collapse problem is that SeqGAN always tends to generate short sequences, since these sequences are easy to be learned well to get higher scores. To enhance the capacity of modeling long-term dependence, hierarchical methods like LeakGAN [Guo *et al.*, 2017] are proposed. LeakGAN has shown promising results in improving long text generation robustness with, however, introducing other kinds of mode collapse according to our experiments.

Meanwhile, inspired by GAN variants like Wasserstein GAN [Arjovsky *et al.*, 2017], diversity-promoting GAN (DP-GAN) [Xu *et al.*, 2018] is proposed to alleviate the collapse mode problem. DPGAN uses an adversarial language model $D_\phi(x_t|s_t)$ to replace the original binary classifier $D_\phi(s)$. Instead of using the saturating binary classification score $D_\phi(s)$, DPGAN designs a hierarchical reward which aims at optimizing the non-saturating objective, namely adversarial $NLL$ estimated by adversarial language model $D_\phi(x_t|s_t)$.

**Re-parametrization**
Besides RL based methods, there are attempts at applying the re-parametrization trick to RNNLMs to bypass the problem of gradient calculation over discrete tokens. As a typical example, GAN for text generation with Gumbel softmax trick [Kusner and Hernández-Lobato, 2016] uses the Gumbel distribution to avoid explicit sampling, making it possible to perform joint training through back-propagation. Adversarial generation of natural language [Rajeswar *et al.*, 2017] introduces randomness through decoding a Gaussian noise in each step to avoid explicit sampling. However, according to our experiments, the reliability of above work implemented through re-parametrization is significantly lower than that of SeqGAN variants. The generated text is seldom readable and of severe mode collapse. Such a kind of problem is revealed from a benchmarking platform, namely Texygen [Zhu *et al.*, 2018], which conducts standard model setting and performs fair comparison among different NTG models.

**Other Methods**
MaskGAN [Fedus *et al.*, 2018] is the first unconditional generative model via sequence-to-sequence (Seq2Seq) learning. Basic version of MaskGAN shares similar ideas with scheduled sampling (SS) [Bengio *et al.*, 2015], yet with adversarial training to address the inconsistency of SS. However, with other useful add-ons like attention mechanism [Bahdanau *et al.*, 2014], MaskGAN has the potential of going far beyond the performance of SS.

Adversarial feature matching for text generation (TextGAN) [Zhang *et al.*, 2017] incorporates adversarial learning through minimizing the reconstruction cost of the adversarial feature estimation of generated text.

### 2.4 On the Limitation of RNNLMs and Beyond

Despite the successful development mentioned above, most current methods have some common properties, which, in some cases, limit their effectiveness.

First, they all follow the RNNLM fomulation that regards the text generation problem as a sequential classification process, each step of which has the form of

$$s_{t-1} \to s_{t-1} \, x_t \, . \quad (15)$$

Note that this Markovian paradigm of production rule indicates that, although RNN variants such as LSTM are proved to be Turing-complete [Siegelmann and Sontag, 1995], language models adopts Eq. (15) can still be regarded as a generalized version of the regular language. As regular language is the simplest one among the four types of languages in the Chomsky Hierarchy, if anything beyond the capacity of regular grammar happens, all the effectiveness of RNNLMs during modeling natural languages will have to rely on generalization ability of RNNs. NNs' generalization ability is not always reliable, making it hard for such models to be well trained. Besides, as the randomness is introduced through sampling from the multinomial distribution in each step [Yu *et al.*, 2017], no explicit latent code can be obtained in this procedure.

On the other hand, some effective architectures like deep convolutional neural network (DCNN or simply CNN) [LeCun *et al.*, 1998] has not been efficiently studied for neural text generation, especially for unconditional text generation. However, as is shown in the paper of WGAN-GP [Gulrajani *et al.*, 2017], it is possible that CNN can be applied to this task via adversarial training.

We particularly look forward to breakthroughs in this direction for a few reasons. First, CNNs make use of the natural locality of languages and at the same time handle long-term dependencies by transforming global dependencies into local dependencies in higher layers of the network. This is like the inverse process of parsing a language through context-free grammar, which has been practically proved useful in many classical NLP tasks. Second, CNNs are parallel-friendly. The training, convergence and inference of CNNs are usually considered to be over ten times faster than RNNs, which is particularly important for practical use.

## 3 Empirical Study

In this section we perform an empirical study of typical neural text generation models. Most of them are evaluated based on Texygen [Zhu *et al.*, 2018], a benchmarking platform particularly for text generation tasks with many well-implemented baseline models and different evaluation metrics. Besides the model that Texygen has already integrated, we also evaluated MaskGAN using the programs provided by the authors[2].

### 3.1 Datasets

**Image COCO**[3] dataset is proposed for image captioning tasks. In the experiment we only use its *image caption annotations*, where we sample 10,000 sentences as training set and another 10,000 as test set. It contains 4,682 distinct words and the maximum length of a sentence is 37. Sentences in this dataset have relatively short and simple patterns.

**EMNLP2017 WMT News**[4] dataset contains news article sentences. Considering the fact that most sentences contain niche words, we just keep sentences containing only the most commonly used 5,700 words. After the pre-process, we choose 200,000 sentences as the training set, 10,000 sentences as the test set. The maximum sentence length is 51 and can be considered as a long text generation dataset.

### 3.2 Metrics

BLEU [Papineni *et al.*, 2002] and $\text{NLL}_{\text{test}}$ [Zhu *et al.*, 2018] are used to evaluate the similarity between documents or generator's capacity to fit real data. Besides, Self-BLEU [Zhu *et al.*, 2018], which calculates the BLEU score between the generated sentences, is used to monitor the severity of mode collapse.

### 3.3 Training Details

**Baselines**
SeqGAN, RankGAN, MaliGAN, TextGAN, LeakGAN and MaskGAN are selected as compared algorithms in the experiment. In addition, the results of standard MLE are also added as a reference.

---

[2] https://github.com/tensorflow/models/tree/master/research/maskgan
[3] http://cocodataset.org/
[4] http://www.statmt.org/wmt17/

Table 1: BLEU score on test data of Image COCO.

| Models | BLEU2 | BLEU3 | BLEU4 | BLEU5 |
|---|---|---|---|---|
| SeqGAN | **0.745** | 0.498 | 0.294 | 0.180 |
| MaliGAN | 0.673 | 0.432 | 0.257 | 0.159 |
| RankGAN | 0.743 | 0.467 | 0.264 | 0.156 |
| LeakGAN | 0.744 | **0.517** | **0.327** | 0.205 |
| MaskGAN | 0.539 | 0.328 | 0.209 | 0.143 |
| TextGAN | 0.593 | 0.463 | 0.277 | **0.207** |
| MLE | 0.731 | 0.497 | 0.305 | 0.189 |

Table 2: BLEU score on test data of EMNLP2017 WMT.

|  | BLEU2 | BLEU3 | BLEU4 | BLEU5 |
|---|---|---|---|---|
| SeqGAN | 0.724 | 0.416 | 0.178 | 0.086 |
| MaliGAN | 0.755 | 0.436 | 0.168 | 0.077 |
| RankGAN | 0.686 | 0.387 | 0.178 | 0.086 |
| LeakGAN | **0.835** | **0.648** | **0.437** | **0.271** |
| MaskGAN | 0.265 | 0.165 | 0.094 | 0.057 |
| TextGAN | 0.205 | 0.173 | 0.153 | 0.133 |
| MLE | 0.771 | 0.481 | 0.249 | 0.133 |

We have also conducted experiments on PG-BLEU, however, in practice, we found that the text generated using models trained by PG-BLEU shares one single pattern, this is because the model is prone to converge on a local optimum in the policy-gradient training process. Thus the results of PG-BLEU are not included in this section.

**Experiment Settings**
In our experiments, all GAN models' parameters are initialized following a standard Gaussian distribution $\mathcal{N}(0, 1)$. Prior to adversarial training, we first pretrain each models' generator and discriminator using the MLE training of 80 epochs respectively, and then we conduct adversarial training of 100 epochs. In the training process of LeakGAN, we use interleaving training scheme proposed by the authors, where 5 MLE epochs will be conducted after every 10 adversarial epochs.

### 3.4 Experiment Results

The BLEU scores on test data of Image COCO and EMNLP2017 WMT are shown in Tables 1 and 2 respectively. LeakGAN shows a great advantage in this metric, especially when the task is long text generation. Among the other models, when generating short text, SeqGAN outperforms other models, while MaliGAN has a slight disadvantage, but it performs quite well on long text generation. MaskGAN and TextGAN do not perform well on this metric.

Figures 2 and 3 depict the $\text{NLL}_{\text{test}}$ curves in the training process. The vertical dashed line represents the end of pretraining process. MaskGAN is excluded in this part since it cannot be evaluated by $\text{NLL}_{\text{test}}$. Since MLE directly optimizes $\text{NLL}_{\text{test}}$ (on the training data), the best scores are always achieved at the end of pretraining except LeakGAN. The $\text{NLL}_{\text{test}}$ loss of LeakGAN tends to converge at an even lower value after it reaches a minima at the end of pretraining, which may be the result of its interleaving training process. TextGAN provides the lowest performance on this metric among all models because its training objective is not about likelihood but the feature distribution distance. The other models share almost identical learning curves during pretraining since they all use standard MLE in this process.

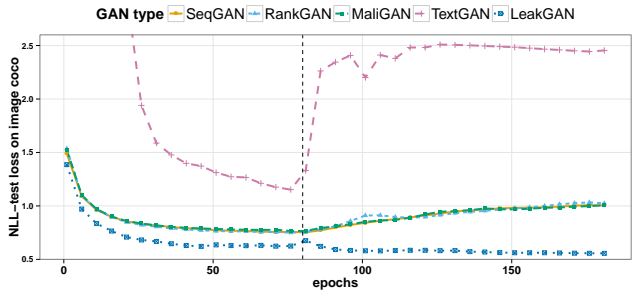

Figure 2: NLL-test loss on Image COCO.

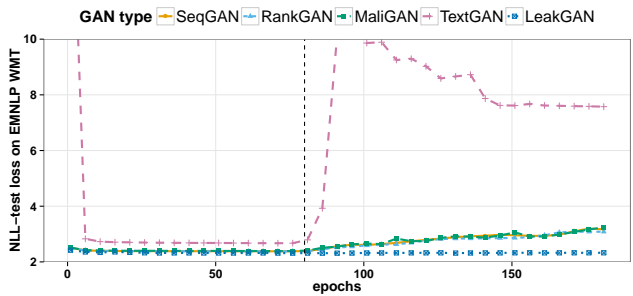

Figure 3: NLL-test loss on EMNLP2017 WMT News.

Table 3: Self-BLEU score on Image COCO.

| Models | BLEU2 | BLEU3 | BLEU4 | BLEU5 |
|---|---|---|---|---|
| SeqGAN | 0.950 | 0.840 | 0.670 | 0.489 |
| MaliGAN | 0.918 | 0.781 | 0.606 | 0.437 |
| RankGAN | 0.959 | 0.882 | 0.762 | 0.618 |
| LeakGAN | 0.934 | 0.818 | 0.663 | 0.510 |
| MaskGAN | **0.752** | **0.516** | **0.378** | **0.293** |
| TextGAN | 0.942 | 0.931 | 0.804 | 0.746 |
| MLE | 0.916 | 0.769 | 0.583 | 0.408 |

Table 4: Self-BLEU score on EMNLP2017 WMT News.

| Models | BLEU2 | BLEU3 | BLEU4 | BLEU5 |
|---|---|---|---|---|
| SeqGAN | 0.907 | 0.704 | 0.463 | 0.265 |
| MaliGAN | 0.909 | 0.718 | 0.470 | 0.252 |
| RankGAN | 0.897 | 0.677 | 0.448 | 0.298 |
| LeakGAN | 0.938 | 0.821 | 0.668 | 0.510 |
| MaskGAN | **0.448** | **0.244** | **0.140** | **0.091** |
| TextGAN | 0.999 | 0.975 | 0.967 | 0.962 |
| MLE | 0.851 | 0.572 | 0.316 | 0.171 |

The Self-BLEU scores are shown in Tables 3 and 4. One can observe that when generating short text, MaskGAN has the least serious mode collapse, as the authors claimed in the paper. On the other hand, TextGAN suffers from very severe mode collapse problem, especially when the training set is the long text. Considering the fact that TextGAN has rather high BLEU score when the n-gram is large, this may result from its mode collapse, which means it'll generate high-frequency phrases in a large amount and repeatedly. Among other models, MaliGAN has the lowest level of mode collapse, but its advantage fades when it comes to generating long texts.

## 4 Conclusion

This paper presents an overview of the classic and recently proposed neural text generation models. The development of RNNLMs are discussed in detail with three training paradigms, namely supervised learning, reinforcement learning and adversarial training. Supervised learning methods with MLE objective are the most widely adopted solution for NTG but they probably cause exposure bias problem. RL-based and adversarial training methods could address exposure bias but usually suffer from gradient vanishing and mode collapse problems. Thus various techniques, including reward rescaling and hierarchical architectures, are proposed to alleviate such problems. This paper also provides a unified view of MLE and RL-based models, which also explains why pretraining with MLE is usually necessary in for RL-based models. The paper also raises a question about whether the effectiveness of RNNLMs is still limited, along with an opinion and corresponding reasons. We hope that this paper could shed a new light on neural text generation landscape and its future research.